\documentclass[10pt,twocolumn,letterpaper]{article}
\usepackage[pagenumbers]{cvpr}  
\usepackage{graphicx}
\usepackage{amsmath}
\usepackage{amssymb}
\usepackage{booktabs}
\usepackage{multirow}
\usepackage{soul}

\usepackage[pagebackref,breaklinks,colorlinks]{hyperref}

\usepackage[capitalize]{cleveref}
\crefname{section}{Sec.}{Secs.}
\Crefname{section}{Section}{Sections}
\Crefname{table}{Table}{Tables}
\crefname{table}{Tab.}{Tabs.}

\begin{document}

\title{Turbo: Informativity-Driven Acceleration Plug-In for Vision-Language Models}

\author{Chen Ju\textsuperscript{1*}, Haicheng Wang\textsuperscript{1*}, Zeqian Li\textsuperscript{1}, Xu Chen\textsuperscript{2}, Zhonghua Zhai\textsuperscript{2}, Weilin Huang\textsuperscript{2}, Shuai Xiao\textsuperscript{2}
\and
\textsuperscript{1}\,{CMIC, Shanghai Jiao Tong University} \;
\textsuperscript{2}\,{TAO Technology, Alibaba Group} \\
{\tt\small \{ju\_chen,\;anakin\_skywalker,\;lzq0103\}@sjtu.edu.cn} \\ 
{\tt\small \{huaisong.cx,\;zhaizhonghua.zzh,\;weilin.hwl,\;shuai.xsh\}@taobao.com}}
\maketitle

\begin{abstract}
Vision-Language Large Models (VLMs) have become primary backbone of AI, due to the impressive performance. However, their expensive computation costs, i.e., throughput and delay, impede potentials in real-world scenarios. To achieve acceleration for VLMs, most existing methods focus on the model perspective: pruning, distillation, quantification, but completely overlook the data-perspective redundancy. To fill the overlook, this paper pioneers the severity of data redundancy, and designs one plug-and-play Turbo module guided by information degree to prune inefficient tokens from visual or textual data. In pursuit of efficiency-performance trade-offs, information degree takes two key factors into consideration: mutual redundancy and semantic value. Concretely, the former evaluates the data duplication between sequential tokens; while the latter evaluates each token by its contribution to the overall semantics. As a result, tokens with high information degree carry less redundancy and stronger semantics. For VLMs' calculation, Turbo works as a user-friendly plug-in that sorts data referring to information degree, utilizing only top-level ones to save costs. Its advantages are multifaceted, e.g., being generally compatible to various VLMs across understanding and generation, simple use without retraining and trivial engineering efforts. On multiple public VLMs benchmarks, we conduct extensive experiments to reveal the gratifying acceleration of Turbo, under negligible performance drop.  
\end{abstract}

\vspace{-0.35cm}
\section{Introduction}  \label{sec:intro}
\vspace{-0.1cm}
Vision-Language Large Models (VLMs)~\cite{Radford21,Jia21,li2022blip,li2023blip,kim2021vilt} have achieved promising progress towards artificial intelligence. Inspired by superior performance and emergent abilities, VLMs are even considered as one of the future trend towards AGI. As some studies~\cite{wei2022emergent,brown2020language} illustrated, VLMs' capability is closely linked to two core factors: model parameters and data quality. Therefore, expanding the model scale and feeding high-quality data to improve the practical performance, has become the community consensus. As expected, VLMs demonstrates superior results on various downstream tasks/domains~\cite{rombach2022high,ju2022prompting,zhou2022learning}, including understanding and generation, with delicate architectures.

\begin{figure}[t]
\begin{center}
\vspace{-0.2cm}
\includegraphics[width=0.475\textwidth] {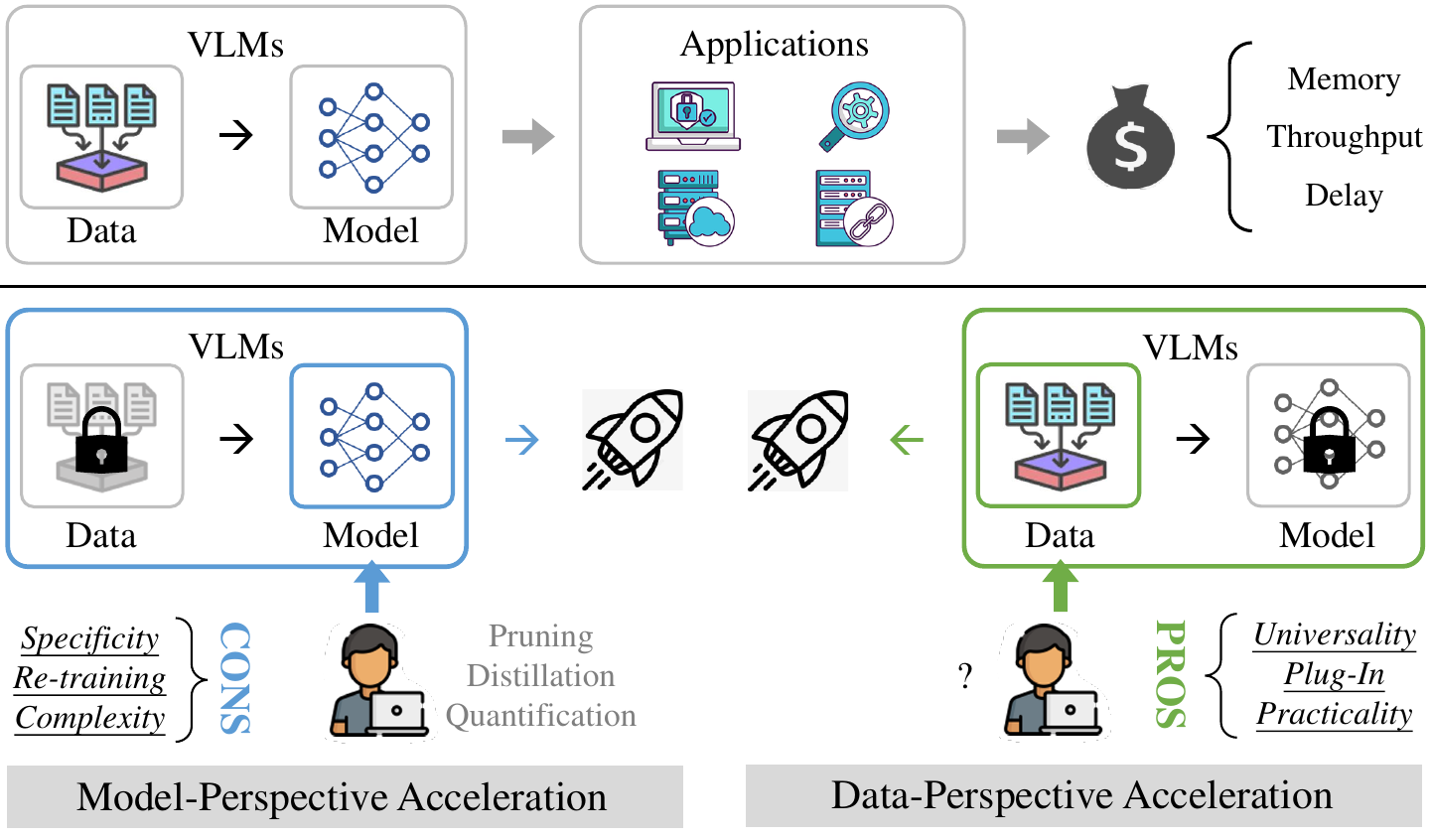}
\end{center}
\vspace{-0.5cm}
\caption{\textbf{Top.} The trouble for VLMs in implementation is the cost issue: memory and throughput. \textbf{Bottom.} Most ideas focus on the model-perspective acceleration, {\em i.e.}, pruning \& quantification. Our Turbo explores data-perspective acceleration by de-redundancy.}
\label{fig:intro}
\vspace{-0.2cm}
\end{figure}

Although it seems that building leading to AI has been completed, there is still a dark cloud hanging over VLMs, making everything a utopia. That is, the cost issue (computational throughput/latency/memory), whether it is training or deployment. Let's imagine, for text-to-image generation, the snail's pace of one picture every 10 seconds is a bottleneck for real-world applications. As a result, reducing costs becomes key to promoting the popularity of VLMs.

To achieve acceleration for VLMs, most existing methods focus on \textbf{one Model-Center Perspective (MCP)}~\cite{wang2022efficientvlm,shen2020q}, {\em e.g.}, pruning, distillation, quantification, as shown in Figure~\ref{fig:intro}. Although effective, they have three following issues: \textbf{Specificity}. MCP is only developed for specific network architectures with poor generalization, which means it's incompatible with various VLMs. For example, MCP acceleration suitable for understanding is usually infeasible for generation. \textbf{Re-training}. In order to maintain high performance, MCP usually requires re-training or fine-tuning VLMs, which inevitably consumes additional computing overheads, {\em i.e.}, being inefficient to apply. \textbf{Complexity}. The development of MCP involves considerable tricks, these trivial matters raises barriers to applications. To sum up, MCP is trapped in one dilemma: easy to hit ceiling, as well as hard to be generalized.

To jump out of the dilemma, this paper raises one novel question about acceleration, from \textbf{one Data-Center Perspective (DCP)}: Is there redundancy in the data? And if so, how high? To take one step answering the question, we first define, then evaluate the informativity of token sequences in each VLMs block. The exploration motivation is that attention-based networks have emerged as a dominant architecture among VLMs, resulting in a quadratic relationship between computation and input sequence length. 
Results show, token redundancy is consistently high in most blocks. Besides, token redundancy gradually increases as blocks deepen. With these observations, we conclude that acceleration through data de-redundancy is promising.

To enable data acceleration, we propose one novel Turbo plug-in module, with the spirit that compressing invalid components while retaining semantic essences. Concretely, we define an information degree, covering two components: mutual redundancy and semantic value. The former evaluates the information duplication between sequential tokens, while the latter focuses on the token’s contribution to sample-wise semantics. For mutual redundancy, the insight is that tokens with dependency tend to have similar informativity, making re-use feasible. For semantic value, the insight is that tokens with core contributions maintain strong performance as principal components. Using information degree of sequential tokens, Turbo naturally sorts data to only leverage the top-level ones for VLMs' calculation, thus saving costs from the source. To sum up, Turbo wins considerable pros. \textbf{Universality}. The accelerated data could directly compatible with on various VLMs, {\em e.g.}, understanding/generation, multi-modality/uni-modality, showing the powerful generalization. \textbf{Plug-and-Play.} Turbo involves no training parameters, which is lightweight to avoid trivial development efforts. \textbf{Practicality}. Turbo is user-friendly with no tedious tricks. Even more valuable, it can complement existing model-perspective acceleration.

Under two types of VLMs, namely, image-align-text understanding and text-to-image generation, we conduct experiments to prove the acceleration generality of Turbo, across several standard datasets. For almost all understanding tasks (retrieval, classification, caption and VQA), Turbo improves throughput by around $2$X with little loss of performance. For most generation tasks (text-to-image, image-to-image), Turbo improves throughput to $1.6$X without compromising quality. We do extensive ablations to reveal the component effectiveness, both quantitatively and qualitatively. Comparing to existing acceleration ideas, Turbo gets better efficiency-performance trade-offs, showing great potential for resource-constrained scenarios.

To sum up, our contributions lie in three folds:

$\bullet$ We pioneer a data-perspective acceleration for VLMs, by token de-redundancy, which is universally adapted to any attention-based understanding and generation models. 

$\bullet$ We design one novel Turbo plug-in to perform data de-redundancy by information degree of redundancy and semantics, getting great efficiency-performance trade-offs.

$\bullet$ We conduct extensive experiments and thorough ablations to reveal the great significance of Turbo acceleration, and our superior results on understanding/generation tasks.

\vspace{-0.2cm}
\section{Related Work}  \label{sec:related work}
\noindent \textbf{Vision-Align-Language Understanding} learns image-text shared embedding spaces, by pre-training with large-scale data. The recent studies mainly divided into three groups, namely, single tower~\cite{chen2020uniter,kim2021vilt}, twin towers~\cite{Radford21,Jia21}, and bridge tower~\cite{li2023blip,li2022blip,zhu2023minigpt}. Structurally, attention-based Transformer has emerged as a dominant architecture, greatly increasing computing overhead while improving performance. They have brought many promising potentials in terms of understanding videos~\cite{ju2021divide,ju2022adaptive,ju2020point}, audio~\cite{liu2023annotation,liu2023audio} and image~\cite{cheng2023mixer,chen2023enhancing}. More specifically, it can be broken down into the following tasks: recognition~\cite{ju2023distilling,zhao2020bottom}, grounding~\cite{ju2023constraint,liu2022exploiting}, segmentation~\cite{ma2023open,yang2023multi}, caption~\cite{mokady2021clipcap,luo2022clip4clip}, retrieval~\cite{ju2022prompting,song2022clip}.

\begin{figure*}[t]
\begin{center}
\includegraphics[width=0.95\textwidth] {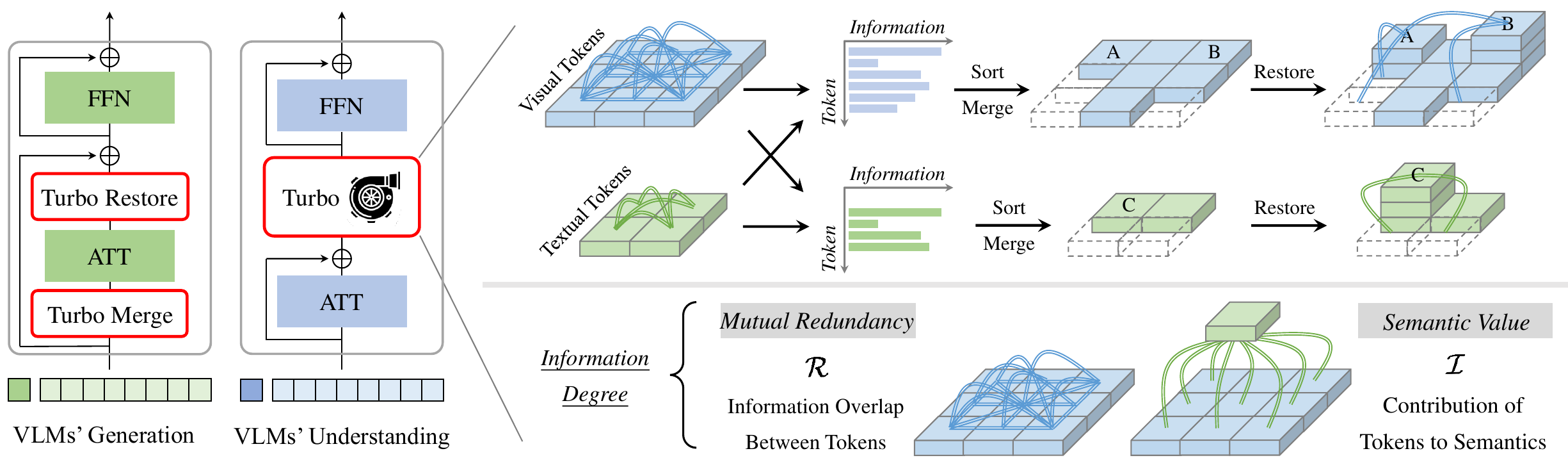}
\end{center}
\vspace{-0.5cm}
\caption{\textbf{Computing Architecture.} 
As a plug-in module, Turbo calculates information degree (mutual redundancy and semantic value) to sort the VLMs’ tokens, and then merges tokens for data reduction. It shows strong generalization across understanding/generation and uni-modality/multi-modality, and is flexible to apply in any block of VLMs to cut computation overheads.}
\label{fig:framework}
\end{figure*}

\vspace{0.05cm}
\noindent \textbf{Vision-Language Generation} aims to build compositional models from text \& noise to pixel-wise vision. Typically, research lines are mainly divided into: Diffusion (SD)~\cite{ho2020denoising,rombach2022high} and DALL-E~\cite{ramesh2022hierarchical,shi2020improving}. Although they have shown promising potentials~\cite{ma2023diffusionseg,ju2023multi}, expensive computing costs seriously damage user experience. For instance, when using SD for one $1024*1024$ image, the inference delay is about $8$ seconds. Structurally, SD’s UNet accounts for most overhead.

\vspace{0.05cm}
\noindent {\bf Acceleration for VLMs.}
To cut high computing costs, extensive work has been proposed to accelerate VLMs, for deployment on resource-limited devices. To sum up, there are four main ideas, namely, knowledge distillation~\cite{wang2022efficientvlm,fang2021compressing}, floating quantification~\cite{shen2020q}, and model pruning~\cite{wang2022efficientvlm,rao2021dynamicvit,shi2023upop} and retraining for data pruning~\cite{jiang2022trips,shi2023crossget,wang2023smarttrim}. Although these methods achieve high sparsity with competitive performance, they all suffer from the model-perspective dilemma, in the following aspects. \textit{Re-training}. To maintain high performance, fine-tuning VLMs is usually inevitable, which is inefficient to consume additional computing overheads. \textit{Specificity}. They are generally developed for specific network architectures, which have poor generalization to compatible with various VLMs. \textit{Complexity}. They could involve many tricks in empirical, these trivial matters raises barriers to application. Overall, comparing to above methods, our Turbo is totally training-free, can be easily plugged in most VLMs with the competitive performance.

Recently, ToMe~\cite{bolya2022token} is been proposed as an acceleration plug-in, but Turbo differs from it in two aspects. \textbf{Universality.} ToMe works only for uni-modal image classification; while Turbo is generally compatible to various VLMs: understanding/generation, multi-modality/uni-modality. \textbf{All-round.} ToMe focuses solely on mutual redundancy; while Turbo combines mutual redundancy and semantic value, to achieve better acceleration-performance trade-offs.

\section{Acceleration for VLMs}   \label{method}
With the goal of alleviating heavy deployment costs from VLMs, we conduct thorough analysis in terms of data; then describe informativity-driven turbo plug-in for acceleration.

\subsection{Preliminaries} 
\label{preliminary}
\noindent {\bf Revisit of Attention.} 
For VLMs, Transformer composed of attention has emerged as a dominant architecture. 
For either text or image data, we generally processes it as the $N$-token 1D sequence $\mathbf{X} \in \mathbb{R}^{N \times D}$, then pass into the attention layer to model sequence relationships globally, where one weighted sum of values based on the affinity over other tokens is calculated. Formally, attention is formulated as: 
\begin{equation} \label{eq:att}
    {\mathrm{Attention}(\mathbf{Q,K,V}) = \mathrm{softmax}(\frac{\mathbf{Q}\mathbf{K}^T}{\sqrt{D}})\mathbf{V},}
\end{equation}

Transformer achieves impressive performance compared to previous architectures, {\em e.g.}, convolution and graph, but at the cost of being more expensive. It's because the complexity of multi-head attention modules (self-attention or cross-attention) scales quadratically with the sequence length $N$, long tokens lead to substantial computational overheads.

With the fundamental understanding, we raise one novel question: Is there any redundancy in the token sequence? If so, we can remove such redundancy from one data perspective, to cut off computational overheads from the source.

\subsection{Multi-Modal Information Redundancy} \label{sec:theory}

\vspace{0.1cm}
\noindent {\bf Motivation \& Insight.} 
Given that most input data are likely to contain superfluous parts, {\em e.g.}, image background, irrelevant objects, we foresee a significant opportunity to further compress the data. In order to accelerate the inference from data perspective, the primary process is to distinguish where the redundancy lies. To analyse it quantitatively, we define token sequence informativity as a measure to quantify the information contained in one token sequence.

Inspired by the informativity idea in information theory, we define $(\Omega, \mathcal{F}, \mathbb{P})$ as the probability space of token sequences, where the triplet is sample space, $\sigma$-algebra and probability measure respectively. For one token sequence $\mathbf{X}=[x_1,x_2,...,x_n]$, we define its informativity as:
\begin{equation}
{\mathcal{I}(\mathbf{X})= -\mathrm{log} \, \mathbb{P}(\mathbf{X}) = -\mathrm{log}\mathbb{P}([x_1,x_2,...,x_n]),}
\end{equation}
By the compound probability formula, we have
\begin{equation}
\mathcal{I}(\mathbf{X}) = -\mathrm{log} \, \mathbb{P}(x_{k, k\in\psi} | x_{i,i\in \{\phi -\psi\}}) \cdot \mathbb{P}(x_{i,i\in \{\phi -\psi\}}),
\end{equation}
where $\phi = \{1,...,n\}$ and $\psi$ is a subset of $\phi$. If we manage to find a subset $\psi$ such that
\begin{equation}
\mathbb{P}(x_{k, k\in\psi} | x_{i,i\in \{\phi -\psi\}}) \approx 1
\end{equation}
then sequence informativity can be approximated by
\begin{equation}
\mathcal{I}(\mathbf{X}) = -\mathrm{log} \, \mathbb{P}(x_{i,i\in \{\phi -\psi\}}).
\end{equation}
This approximation demonstrates that if we manage to find a subset of token sequence $\{x_{k, k\in\psi}\}$ which have a total dependency on its complementary set $\{x_{i,i\in \{\phi -\psi\}}\}$, then we could aggregate $\{x_{k, k\in\psi}\}$ into its complementary set for more efficient calculation without much information loss.

\vspace{0.1cm}
\noindent {\bf Mutual Redundancy.} 
According to the above analysis, we would like to find the dependencies between tokens of the given sequence. MAE~\cite{he2022masked} shows that there exists considerable redundancies in image tokens, such that only $25$\% tokens can almost restore the whole sequence, thus a such subset $\psi$ is verified to exist. However, it's nontrivial to find a proper subset, for the token dependencies is unknown. 

One intuition is that tokens with mutual dependency tend to have similar representations, {\em e.g.}, we could easily restore the texture of a shirt by just a few patches from it. What's more, notice that tokens with high similarity have analogical contributions in the attention calculation. As some studies~\cite{bolya2022token} have shown, token aggregation by mutual similarity is able to preserve key information. So by pairing up tokens with similar representations, we manage to construct a subset $\{x_{k, k\in\psi}\}$ with a high dependency on $\{x_{i,i\in \{\phi -\psi\}}\}$.

\vspace{0.1cm}
\noindent {\bf Semantic Value.} 
Mutual redundancy for data pruning can already get relatively good results on simple tasks, {\em i.e.}, image classification. However, in this process, each token is treated equally, thus its contribution to semantic categories are neglected. This will result in an early merging for tokens with important semantics, causing a drastic performance drop on fine-grained, information-demanding tasks, {\em i.e.}, multi-modal understanding and generation.

To merge tokens with a preference from trivial background to significant foreground, we leverage the guidance of semantic value hidden inside the network. We consider $\mathit{cls}$ token as the semantic enrichment. For $\mathit{cls}$ token $\mathbf{Y}$, we can calculate its informativity as:
\begin{equation}
\mathcal{I}(\mathbf{Y})= -\mathrm{log} \, \mathbb{P}(\mathbf{Y}) = -\mathrm{log} \, \mathbb{P}(\mathbf{A} \cdot \mathbf{V}),
\end{equation}
where $\mathbf{V}$ is the affine transformation of token sequence $\mathbf{X}$, and $\mathbf{A}$ refers to the attention map for the $\mathit{cls}$ token: 
\begin{equation} \label{eq:44}
    {
    \mathbf{A} = \mathrm{softmax}(\frac{\mathbf{Q}_{\mathrm{cls}} \, \mathbf{K}^T}{\sqrt{D}}) \in \mathbb{R}^{1 \times n}, \sum_{i=1}^{n} \mathbf{A}_{1,i} = 1.
    }
\end{equation}
We define the \textbf{semantic value} of $i$-th token as its attention weight $A_{1,i}$. 
As we have no access to the probability distribution of $\mathbf{V}$, here we adopt one meandering way to approximate the solution. 
Inspired by vector quantization~\cite{esser2021taming,van2017neural}, one common sense can be summarized as follows. In the context of neighbourhood, continuous variables can be approximately represented by discrete quantities, which illustrate the fact that if we place the perspective in a local neighborhood, vectors that are closer in distance tend to possess similar semantic meanings. Thus we make a first-order approximation that under the condition of neighbourhood near semantic-rich $y_1\in R^n$ with range $\epsilon>0$, for all $y_2\in R^n$ satisfying $\Vert y_1 - y_2 \Vert_2 < \epsilon$, there exists an $\beta \in R_+$ so that
\begin{equation} \label{eq:55}
    {
    \Vert \mathcal{I}(y_1) - \mathcal{I}(y_2) \Vert \ \leq \  \beta \Vert y_1 - y_2  \Vert_2. 
    }
\end{equation}
Denote $\mathbf{Y}'$ the $\mathit{cls}$ token after pruning the tokens $\{x_{j, j\in \psi'}\}$, with $\psi' \subseteq \{1,...,n\}$:
\begin{equation}
\mathbf{Y}'=\sum_{i=1}^n \mathbf{A}_i \mathbf{V}_i - \sum_{j \in \psi'} \mathbf{A}_j \mathbf{V}_j = \mathbf{A'}\cdot \mathbf{V'},
\end{equation}
Replacing $y_1$ by the original $\mathit{cls}$ token $\mathbf{Y}$ and $y_2$ by the pruned $\mathit{cls}$ token $\mathbf{Y}'$ in Eq.~\ref{eq:55}:
\begin{equation}
    {
    \Vert \mathcal{I}(\mathbf{A}\cdot \mathbf{V}) - \mathcal{I}(\mathbf{A'}\cdot \mathbf{V'}) \Vert \ \leq \ \beta \Vert \mathbf{A}\cdot \mathbf{V} - \mathbf{A'}\cdot \mathbf{V'}  \Vert_2. 
    }
\end{equation}
Suppose we have
\begin{equation}
\Vert \sum_{j \in \psi'} \mathbf{A}_j \mathbf{V}_j \Vert_2 \ll \frac{1}{\beta},
\end{equation}
With the prior assumption in Eq.~\ref{eq:55}, we can approximate the informativity of $\mathit{cls}$ token $\mathcal{I}(\mathbf{Y})$ as $\mathcal{I}(\mathbf{Y}')$ because 
\begin{equation} \label{eq:12}
    {
    \Vert \mathcal{I}(\mathbf{Y}) - \mathcal{I}(\mathbf{Y}') \Vert \leq \ \beta \Vert \sum_{j \in \psi'} \mathbf{A}_j \mathbf{V}_j \Vert_2 \ll 1.
    }
\end{equation}
Such theoretical analysis indicates that, pruning/merging tokens with sufficiently small semantic values, will barely affect the informativity of $\mathit{cls}$ token.

Semantic value captures token importance/relevance to the overall semantics. We add it into the merging strategy, giving tokens with high semantics less weight during the reduction. This ensures that tokens with high semantic contributions are more likely to be retained, even if their mutual redundancy is high. Figure~\ref{fig:abla} and Section~\ref{sec:abla} implies the mentioned properties of semantic value.

\vspace{0.1cm}
\noindent \underline{\textit{Remark.}}
For uni-modal tasks, we adopt visual \textit{cls} token as semantic guidance; while for multi-modal tasks, we use visual or textual \textit{cls} token depending on model architectures.

\begin{figure}[t]
\begin{center}
\includegraphics[width=0.43\textwidth] {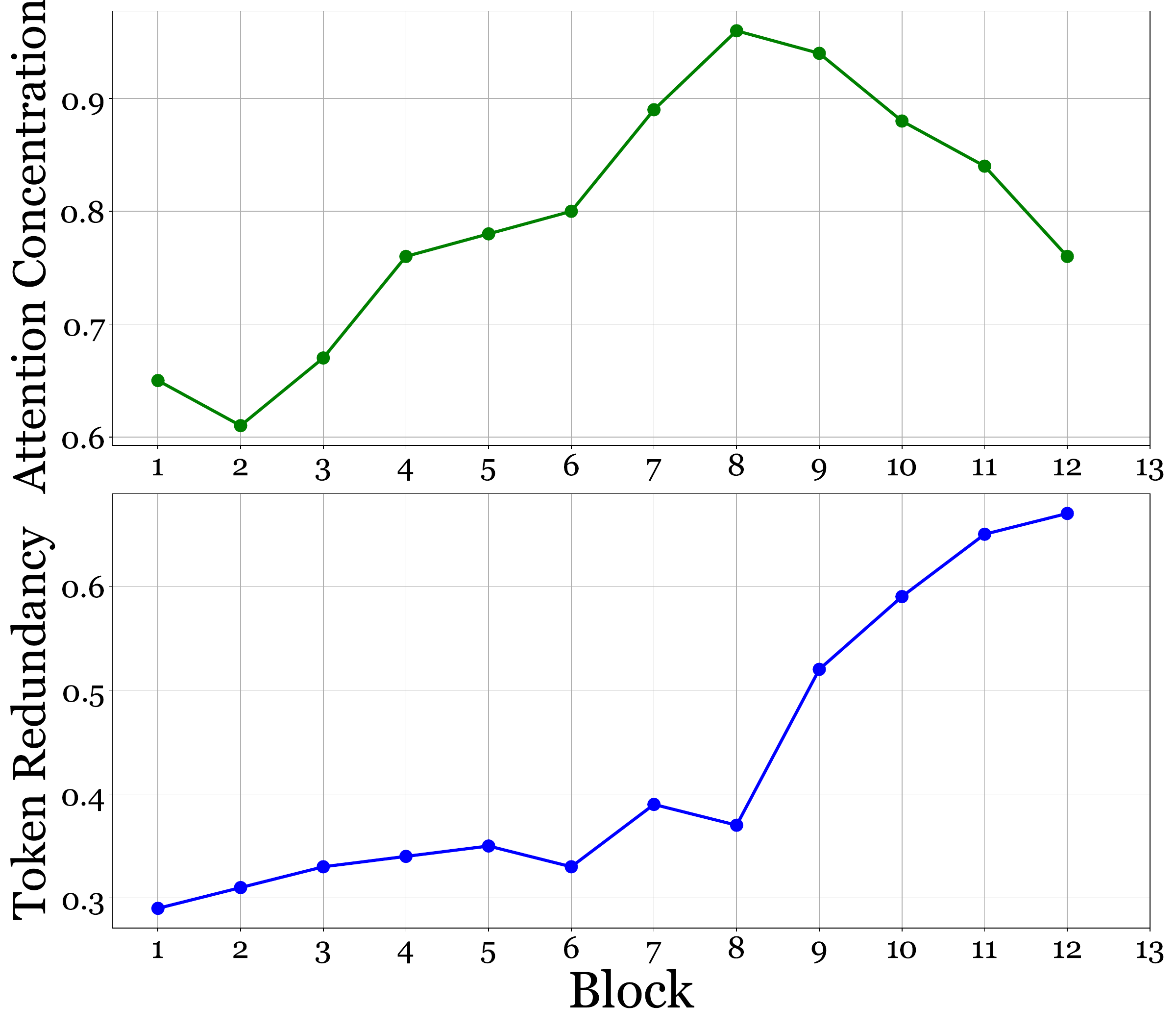}
\end{center}
\vspace{-0.6cm}
\caption{\textbf{Empirical Evaluation of Redundancy.} We experiment on BLIP image encoder fine-tuned for multi-modal retrieval. The results support our argument for potential redundancies in data.}
\label{fig:evaluation}
\end{figure}

\subsection{Empirical Evaluation of Redundancy} 
As proposed above, we identify mutual redundancy and semantic value as the two main sources of redundancy in the inference process. To quantitatively measure these redundancies in the output token sequence $\mathbf{X}=[\mathbf{x}_1,\mathbf{x}_2,...,\mathbf{x}_n]$ of one attention layer, we first define evaluation metrics for mutual redundancy, named as token redundancy, {\em i.e.}, measure cosine similarities $\mathcal{S}$ between tokens:
\begin{equation} \label{eq:simi}
    {\mathcal{S}=\frac{2}{n*(n-1)}\sum_{i=1}^n\sum_{j=i+1}^n\frac{\mathbf{x}_i\cdot \mathbf{x}_j}{\left \| \mathbf{x}_i \right \|_2 \left \| \mathbf{x}_j \right \|_2},}
\end{equation}
where $\left \| \cdot \right \|_2$ is the L2 norm and $\cdot$ is the dot product. 

Then we define the attention concentration $\mathcal{C}$ as the proportion of semantic values in the top $\frac{1}{4}$ tokens with highest values:
\begin{equation} \label{eq:conc}
    {\mathcal{C}=\sum_{i=1}^{\lfloor \frac{n}{4} \lfloor}\mathbf{A}_{1,i},}
\end{equation}
where $\mathbf{A}$ is the semantic value defined in Eq.~\ref{eq:44}. We use this measure to verify the long-tail distribution of semantic values. Larger $\mathcal{C}$ implies a higher degree of concentration on few specific tokens, encouraging safe reduction to insignificant trailing tokens without losing much information.

To view underlying redundancies in VLMs, we experiment on image encoder of BLIP~\cite{li2022blip}, with the parameters fine-tuned on COCO retrieval datasets. Figure~\ref{fig:evaluation} counts token redundancy and attention concentration. By analyzing results, we empirically draw the following conclusions:

\vspace{0.05cm}
\noindent $\bullet$ \textbf{Basis}: token redundancy and attention concentration are consistently high in most blocks, achieving $95\%$ in intermediate layers, {\em i.e.}, the left $\frac{3}{4}$ tokens contributes only $5\%$.

\vspace{0.05cm}
\noindent $\bullet$ \textbf{Trend}: token redundancy gradually increases as blocks deepen, implying a tendency of clustering done by the network. Attention concentration presents a bottleneck trend, with the maximum value reached at the intermediate layer. 

These above observations support the fact that tokens in the attention layer possess high mutual redundancy and concentrated semantic values, thus acceleration by data de-redundancy is promising. We next achieve the goal by one novel Turbo plug-in.

\subsection{Plug-and-Play Turbo}  
\noindent {\bf Strategy.} 
Information degree $\mathcal{E}$ is proposed to balance mutual redundancy $\mathcal{R}$ and semantic value $\mathcal{I}$, we explore two fusion strategies: weighted averaging or coupled product. 
\begin{equation}  \label{eq:balance}
    {\mathcal{E} = \mathcal{R}-\alpha\mathcal{I}, \quad \mathcal{E} = \frac{\mathcal{R}}{\mathcal{I}}.
    }
\end{equation}
Table~\ref{tab:degree} makes detailed comparisons, and shows the significance of trade-offs between $\mathcal{R}$ and $\mathcal{I}$.

With information degree $\mathcal{E}$, we naturally propose a Turbo module to accelerate by data de-redundancy. Since the data organization of most VLMs is attention-based sequences, Turbo could perform on any block with great flexibility.

\vspace{0.1cm}
\noindent {\bf Role.} Structurally speaking, Turbo does behave in differentiated forms for generation and understanding. 

For high-level semantic tasks (such as, classification, retrieval, caption, VQA), the role of Turbo is to calculate the information degree for $N$ input tokens $\mathbf{X} \in \mathbb{R}^{N \times D}$, and then sort them to merge the top-level ones to the rest $\mathbf{X}{'}$. Formally: 
\begin{equation}  \label{eq:highlevel}
    {\mathbf{X}{'} =\Phi_{\mathrm{tub}}^{\mathrm{high}}(\mathbf{X}, \; \Upsilon) \in \mathbb{R}^{N\cdot(1-\Upsilon) \times D}
    }
\end{equation} 
where $\Upsilon$ is the token drop ratio to all sequential tokens and $\Phi_{\mathrm{tub}}^{\mathrm{high}}$ is the merging operation for high-level tasks. For parallel computing, we set $\Upsilon$ to one constant in each batch.

For low-level generative tasks (text-to-image or image-to-image composition), the fine-grained pixel requirements make simple merging of redundant tokens infeasible. We hence divide Turbo into two modules, making its role become \textit{merge \& restore}. Specifically, before VLMs' block, Turbo calculates and records information degree between all tokens, then removes redundant tokens using a merge module $\Psi_{\mathrm{meg}}$; while after VLMs' block, Turbo restores the merged tokens with reference to information degree, using one inverse-sampling module $\Psi_{\mathrm{inv}}$. 
Formally, 
\begin{equation}  \label{eq:lowlevwl}
    {\mathbf{X}{'} = \Phi_{\mathrm{tub}}^{\mathrm{low}}(\mathbf{X}, \Upsilon) = 
    \Psi_{\mathrm{inv}}(\Psi_{\mathrm{meg}}(\mathbf{X}, \Upsilon)) \in \mathbb{R}^{N \times D},
    }
\end{equation}
Such Turbo greatly reduces computing overhead for VLMs' blocks, while ensuring pixel-level generation. Note that, in above process, we employ bipartite soft matching~\cite{bosek2014online} to calculate information degree, resulting in negligible additional FLOPs. Besides, Turbo allows the bulk of computing to be done in parallel, making it friendly to modern GPU devices.

\vspace{0.1cm}
\noindent \underline{\textit{Remark.}} 
Comparing to most acceleration ideas, Turbo wins many pros. 
\textbf{Universality}. It performs data de-redundancy, {\em i.e.}, reducing computing from the source of the AI system. The resultant data can be generally adopted on various VLMs, {\em e.g.}, understanding/generation, multi-modality/uni-modality, showing powerful compatibility. 
\textbf{Practicality}. It fully considers the potential data-repetition guided by semantics, accelerating with little performance loss. Besides, it's user-friend without cumbersome tricks. 
\textbf{Plug-and-Play.} It works as plug-in without additional parameters for re-training, and is plain to avoid trivial development efforts.

\section{Experiments}
We here conduct experiments on tasks of understanding and generation across several public benchmarks.

\subsection{Implementation}
\noindent \textbf{Understanding Benchmarks.} 
We evaluate Turbo on uni-modal Image Classification, and multi-modal tasks like Cross-modal Retrieval, Natural Language for Visual Reasoning, Visual Question Answering and Image Captioning. We conduct image classification experiments on AugReg~\cite{steiner2021train} and SWAG~\cite{singh2022revisiting} models for Imagenet-1k dataset and all cross-modal tasks on BLIP~\cite{li2022blip} and BLIP2~\cite{li2023blip} using datasets Flickr30k, COCO and NLVR2.

\vspace{0.1cm}
\noindent \textbf{Generation Benchmarks.} 
We evaluate on Stable Diffusion 1.5~\cite{rombach2022high} by generating $2000$ images, each resolution is $512*512$. The text classes used are from ImageNet-1k. We use FID scores to metric the generation quality.

The above experiments are conducted on one single 3090 GPU. For more details towards Turbo on different benchmarks, please refer to the supplementary materials.

\subsection{Comparison with State-of-the-art}
We carry out full comparisons with SOTA on both uni-modal and multi-modal VLMs, to prove the effectiveness.

\begin{table}[t]
\small
\centering
\begin{tabular}{c|cc|c|cc}
\toprule
\multirow{2}{*}{PlugIn} & \multicolumn{2}{c|}{Performance} & \multirow{2}{*}{\begin{tabular}[c]{@{}c@{}}Ratio\\ $\Upsilon$ \end{tabular}} & \multicolumn{2}{c}{Acceleration} \\ \cline{2-3} \cline{5-6} 
                        & Dev        & Test        &                        & FLOPs       & Throughput         \\ \hline \hline
-                       & 82.5          & 83.3           & -                      & 132.5       & 117.5              \\ \hline
UPop                   & 80.3          & 81.1            & -                      & 89.4 (-0.32×)       & 138.5 (1.18×)      \\
ToMe                    & 79.3          & 79.5           & 35                     & 77.2 (-0.42×)        & 182.9 (1.56×)      \\ \hline \hline
\multirow{2}{*}{Turbo}  & \textbf{81.4}          & \textbf{82.2}           & 35                     & 77.2 (-0.42×)       & 182.8 (1.56×)      \\
                        & 80.5           & 81.5           & 45                     & 62.2 (-0.53×)        & \textbf{224.2 (1.91×)} 
                        \\ \hline \hline
UPop*                       & 79.0          & 79.8           & 35           & \textbf{54.7 (-0.59×)}      & 188.2 (1.60×)
                        \\ \bottomrule
\end{tabular}
\vspace{-0.2cm}
\caption{\textbf{NLVR Acceleration for BLIP on the NLVR2 dataset.} Ratio denotes the number of reduced tokens per layer and - denotes the original accuracy of BLIP model on NLVR. In UPop*, we apply Turbo to the model pruned by UPop, proving that our method is perpendicular to those model-oriented methods. Our method surpasses the other approaches by a large margin both on inference speed and accuracy. }
% \vspace{-0.3cm}
\label{tab:nlvr}
\end{table}

\begin{table}[t]
\small
\setlength\tabcolsep{5pt}
\centering
\begin{tabular}{c|cc|c|cc}
\toprule
\multirow{2}{*}{PlugIn} & \multicolumn{2}{c|}{Performance} & \multirow{2}{*}{\begin{tabular}[c]{@{}c@{}}Ratio\\ $\Upsilon$ \end{tabular}} & \multicolumn{2}{c}{Acceleration} \\ \cline{2-3} \cline{5-6} 
                        & Dev        & Std       &                        & FLOPs           & Throughput     \\ \hline \hline
-                       & 77.4            & 77.5           & -                      & 92.1            & 148.1          \\ \hline
UPop                    & 76.3            & 76.3           & -                      & 65.2 (-0.30×)   & 167.8 (1.13×)  \\
ToMe                    & 74.8            & 74.8           & 40                     & 65.7 (-0.29×)   & 184.8 (1.25×)  \\ \hline \hline
\multirow{2}{*}{Turbo}  & \textbf{76.8}            & \textbf{76.9}           & 40                     & 65.7 (-0.29×)   & 184.6 (1.25×)  \\
                        & 76.6            & 76.7           & 60                     & 52.5 (-0.43×)   & \textbf{232.7 (1.57×)}  
                                                \\ \hline \hline
UPop*              & 75.4          & 75.4           & 40          & \textbf{44.8 (-0.51×)}      & 206.8 (1.40×)
                        \\ \bottomrule
\end{tabular}
\vspace{-0.2cm}
\caption{\textbf{VQA Acceleration for BLIP on the VQAv2 dataset.} Since we adopt large models for VQA task, the drop ratio is higher than base models. By applying $ratio=60$, we achieve the best trade-off between performance reservation and acceleration.}
\label{tab:vqa}
\end{table}

\vspace{0.1cm}
\noindent \textbf{Multi-Modal VLMs for Understanding.} 
A thorough experiment on a wide range of multi-modal tasks on foundation models BLIP~\cite{li2022blip}, and BLIP2~\cite{li2023blip} are done to test the effectiveness of our method. Part of the results of understanding tasks are shown in Tables~\ref{tab:nlvr},~\ref{tab:vqa},~\ref{tab:caption} and~\ref{tab:retrieval}. We compare our method Turbo to the baseline~\cite{bolya2022token} and one model-pruned based approach UPop~\cite{shi2023upop}. Overall, Turbo achieves the best trade-off on acceleration rate and the performance preservation on all benchmarks. We exceed the baseline by a large margin on almost all tasks, meanwhile achieving better or competitive performance to training-required method UPop with less inference time. Compared to model-pruning methods that require further training specially designed for different models, Turbo needs no extra training and can play simple plug-in for various models. What's more, due to the considerable cost of memory access time, FLOPs is not inversely proportional to the actual throughput. This results in a poor acceleration performance for model-pruning method on real scenarios, because these methods can't reduce the memory access time of data. In contrast, by directly pruning data, our Turbo saves both memory access time and calculation amount.

\vspace{0.05cm}
\noindent \textbf{NLVR}: Table~\ref{tab:nlvr} demonstrates the result on NLVR task under different drop ratios $\Upsilon$. Turbo performs the best even with an acceleration of $1.91×$ compared to $1.56×$ for the baseline and $1.18×$ for UPop. We also apply Turbo Plug-in into the model pruned by UPop (Note as Upop*). The results show that Turbo is perpendicular to the model-perspective acceleration, further reflecting the universality of Turbo.

\vspace{0.05cm}
\noindent \textbf{VQA}: Table~\ref{tab:vqa} evaluates on the VQA task for BLIP. Turbo again exceeds the other methods either on performance or acceleration even with large drop ratios. 

\begin{table}[t]
\small
\setlength\tabcolsep{4pt}
\centering
\begin{tabular}{c|c|cc|cc}
\toprule
\multirow{2}{*}{Model} & \multirow{2}{*}{PlugIn} & \multicolumn{2}{c|}{Performance} &  \multicolumn{2}{c}{Acceleration} \\ \cline{3-4} \cline{5-6} 
          &               & B@4        & CIDEr                & FLOPs       & Throughput      \\   
\hline \hline
\multirow{5}{*}{BLIP}  & -      & 39.7 & 133.3 & 330.2  & 34.2         \\
                       & UPop   & \textbf{38.6} & 128.9 & 137.9 (-0.58×)  & 56.6 (1.65×) \\
                       & ToMe   & 35.5 & 120.9 & 134.2 (-0.60×)  & 67.9 (1.99×) \\
                       & Turbo  & 38.2 & \textbf{130.0} & 134.2 (-0.60×)  & 67.6 (1.98×) \\
                       & UPop* & 37.3 & 126.0 & \textbf{60.8 (-0.82×)}  & \textbf{74.8 (2.19×)} \\ \hline \hline
\multirow{3}{*}{BLIP2} & -      & 42.7 & 145.7 & 1379.2 & 29.5         \\
                       & ToMe   & 40.8 & 137.7 & 1029.5 (-0.25×) & 51.1 (1.73×) \\
                       & Turbo  & \textbf{42.1} & \textbf{142.2} & \textbf{989.7 (-0.29×)}  & \textbf{50.9 (1.72×)} \\ 
\bottomrule
\end{tabular}
\vspace{-0.15cm}
\caption{\textbf{Caption Acceleration on COCO.} The drop ratio is 16 and (30, 40) for Turbo on BLIP2 and BLIP respectively. CIDEr and B@4 are two standards for captioning evaluation. Turbo attains competitive performance while largely enhance throughput.
} 
\label{tab:caption}
\end{table}

\vspace{0.05cm}
\noindent \textbf{Image Captioning}: We demonstrate the result of image captioning on both BLIP and BLIP2 in Table~\ref{tab:caption}. As a difficult multi-modal task, image captioning requires fine-grained information. The baseline ToMe performs poorly on this task, dropping from 133.3 to 120.9 on CIDEr, implying a non-negligible loss of information during token merging. However, Turbo can retain much better performance to 130.0 with the same FLOPs and throughput, proving that our method can preserve more fine-grained information by encouraging aggregation of background rather than foreground. We also show that turbo could further accompany model-pruning methods to get even faster inference.

\vspace{0.05cm}
\noindent \textbf{Multi-modal Retrieval}: Along with the image captioning, multi-modal retrieval often deal with large amount of data, {\em i.e.} millions or even billions of image-text pairs in real scenarios, so there is an urgent demand for inference speed-up. Table~\ref{tab:retrieval} demonstrates that, Turbo exceeds the baseline and model-pruning methods by a large margin on several benchmarks, {\em i.e.} $4.3\%$ for image-text and $6.7\%$ for text-image and on BLIP Flickr30k compared to UPop. With such an acceptable accuracy drop, our method is expecting to make large-scale retrieval and captioning more efficiently.

\begin{table*}[t]
\small
\centering
\begin{tabular}{c|c|c|cc|cc|cc}
\toprule
\multirow{2}{*}{Model} & \multirow{2}{*}{Dataset}   & \multirow{2}{*}{PlugIn} & \multicolumn{2}{c|}{Image-To-Text} & \multicolumn{2}{c|}{Text-To-Image} & \multicolumn{2}{c}{Acceleration} \\ \cline{4-9}
                       &                            &                         & R@1            & R@5            & R@1            & R@5            & FLOPs           & Throughput     \\ \hline \hline 
\multirow{6}{*}{BLIP2~\cite{li2023blip}} & \multirow{3}{*}{Flickr30k} & -                       & 97.6           & 100.0          & 89.7           & 98.1           & 717.5                  & 10.7                        \\ \cline{3-9} 
                       &                            & ToMe~\cite{bolya2022token}                    & 96.4           & 100.0          & 86.5           & 97.2           & 376.1 (-0.47X)         & \textbf{19.9 (+1.86X)}               \\  
                       &                            & Turbo                   & \textbf{96.7}$_{+0.3}$    & \textbf{100.0}$_{+0.0}$    & \textbf{87.9}$_{+1.4}$     & \textbf{97.4}$_{+0.2}$     & \textbf{376.0 (-0.47X)}         & 19.7 (+1.85X)               \\ \cline{2-9} 
                       & \multirow{3}{*}{COCO}      & -                       & 85.4           & 97.0           & 68.3           & 87.7           & 717.5                  & 10.8                        \\  \cline{3-9}
                       &                            & ToMe~\cite{bolya2022token}                    & 83.6           & 95.8           & 66.4           & 86.7           & 396.5 (-0.45X)         & \textbf{18.7 (+1.73X)}               \\
                       &                            & Turbo                   & \textbf{84.2}$_{+0.6}$     & \textbf{96.2}$_{+0.4}$     & \textbf{67.1}$_{+0.7}$     & \textbf{87.1}$_{+0.4}$     & \textbf{396.5 (-0.45X)}         & 18.6 (+1.72X)               \\ \hline \hline
\multirow{10}{*}{BLIP~\cite{li2022blip}} & \multirow{5}{*}{Flickr30k} & -                       & 97.2           & 99.9           & 87.3           & 97.6           & 55.5                   & 281.0                       \\ \cline{3-9} 
                       &                            & ToMe~\cite{bolya2022token}                    & 93.7           & 99.5           & 80.1           & 95.5           & 37.0 (-0.33X)          & 375.6 (+1.34X)              \\ 
                       &                            & UPop~\cite{shi2023upop}                    & 92.5           & 99.0           & 78.4           & 94.5           & 39.1 (-0.29X)          & 322.2 (+1.15X)              \\  \cline{3-9} 
                       &                            & \multirow{2}{*}{Turbo}                  & \textbf{96.8}$_{+3.1}$           & \textbf{99.8}$_{+0.3}$           & \textbf{85.1}$_{+5.0}$           & \textbf{96.8}$_{+1.3}$           & 37.0 (-0.33X)          & 375.0 (+1.33X)              \\
                       &                            &                    & 94.2           & 99.4           & 82.9           & 96.0           & \textbf{31.1 (-0.44X)}          & \textbf{449.2 (+1.60X)}              \\ \cline{2-9} 
                       & \multirow{5}{*}{COCO}      & -                       & 81.9           & 95.4           & 64.3           & 86.1           & 55.5                   & 285.0                       \\ \cline{3-9} 
                       &                            & ToMe~\cite{bolya2022token}                    & 74.2           & 92.1           & 56.5           & 80.4           & 36.8 (-0.34X)          & 381.3 (+1.34X)              \\
                       &                            & UPop~\cite{shi2023upop}                    & 77.4           & 93.4           & 59.8           & 83.1           & 39.1 (-0.29X)          & 323.2 (+1.13X)              \\ \cline{3-9} 
                       &                            & \multirow{2}{*}{Turbo}                   & \textbf{78.8}$_{+4.6}$           & \textbf{94.7}$_{+2.6}$           & \textbf{61.3}$_{+4.8}$           & \textbf{84.2}$_{+3.8}$           & 36.8 (-0.34X)          & 380.5 (+1.34X)              \\
                       &                            &                    & 77.5           & 93.8           & 60.5           & 83.7           & \textbf{34.1 (-0.39X)}          & \textbf{424.3 (+1.49X)}
                       \\ \bottomrule
\end{tabular}
\vspace{-0.15cm}
\caption{\textbf{Retrieval Acceleration on BLIP/BLIP2 across Flickr30K/COCO datasets.} The drop ratio is 16 and (30,40) for Turbo on BLIP2 and BLIP respectively. Turbo achieves superior results compared to other methods, with a performance drop of merely $0.4\%$ on BLIP image-text retrieval, while other two approaches drops $3.5\%$ and $4.7\%$ respectively.}
\label{tab:retrieval}
\end{table*}

\begin{table}[t]
\small
\centering
\begin{tabular}{c|c|cc|cc}
\toprule
\multirow{2}{*}{PlugIn} & \multirow{2}{*}{\begin{tabular}[c]{@{}c@{}}Ratio\\ $\Upsilon$ \end{tabular}} & \multicolumn{2}{c|}{Text-To-Image} & \multicolumn{2}{c}{Image-To-Image} \\ \cline{3-6} 
                        &                                                                    & FID           & Throughput         & FID           & Throughput         \\ \hline  \hline
-                       & -                                                                  & 32.12         & 0.32               & 30.04         & 0.29               \\ \hline  \hline
\multirow{2}{*}{ToMe}   & 10                                                                 & 32.80         & 0.39               & 30.69         & 0.37               \\
                        & 20                                                                 & 32.86         & 0.44               & 30.75         & 0.40               \\ \hline  \hline
\multirow{2}{*}{Turbo}  & 20                                                                 & 32.63         & 0.43               & 30.48         & 0.42               \\
                        & 30                                                                 & 32.77         & 0.50               & 30.56         & 0.49               \\ \bottomrule
\end{tabular}
\vspace{-0.2cm}
\caption{\textbf{Generation Acceleration for Stable Diffusion.} Turbo wins higher generation quality (FID) or faster throughput acceleration, regardless of text-to-image or image-to-image tasks.}
\label{tab:sdgene}
\end{table}

\vspace{0.1cm}
\noindent \textbf{Multi-Modal Generation VLMs.} 
Our Turbo is also suitable for fine-grained visual generation. Table~\ref{tab:sdgene} compares the performance-acceleration across text-to-image, image-to-image tasks. 
Comparing to the competitor, Turbo obtains better generation quality, {\em i.e.}, lower FID scores, with a an even better acceleration. While achieving similar generation results, Turbo brings significant throughput gains over vanilla stable diffusion models. Limited by paper pages, we refer the readers to the supplementary materials for detailed visualizations and ablation studies.

\vspace{0.1cm}
\noindent \textbf{Uni-Modal Foundation Models.} 
We perform uni-modal experiments on ImageNet-1k with two different training plans: AugReg~\cite{steiner2021train} and SWAG~\cite{singh2022revisiting}. Table~\ref{tab:uni} compares between our method and baseline ToMe~\cite{bolya2022token}. Our Turbo surpasses ToMe on models of all sizes and training plans, while keeping the same acceleration rate. Note that image classification is rather simple comparing to multi-modal understanding tasks, so the performance drop is relatively small for the baseline. Nevertheless, Turbo achieves a thorough improvement on all uni-modal benchmarks.

\begin{table}[t]
\small
\centering
\begin{tabular}{c|c|c|c|c}
\toprule
Model                 & Backbone                  & PlugIn  & Acc  & Acceleration           \\ \hline \hline
\multirow{6}{*}{ViT~\cite{steiner2021train}}  & \multirow{2}{*}{ViT-S/16} & ToMe     & 81.4/79.3 & \multirow{2}{*}{1.53×} \\                             \cline{3-4}
                      &                           & Turbo    & 81.4/\textbf{79.9} &                        \\ \cline{2-5} 
                      & \multirow{2}{*}{ViT-B/16} & ToMe     & 84.5/82.6 & \multirow{2}{*}{1.62×} \\ \cline{3-4}
                      &                           & Turbo    & 84.5/\textbf{83.1} &                        \\ \cline{2-5} 
                      & \multirow{2}{*}{ViT-L/16} & ToMe     & 85.8/84.2 & \multirow{2}{*}{1.71×} \\ \cline{3-4}
                      &                           & Turbo    & 85.8/\textbf{84.6} &                        \\ \hline \hline
\multirow{6}{*}{SWAG~\cite{singh2022revisiting}} & \multirow{2}{*}{ViT-B/16} & ToMe     & 85.3/84.5 & \multirow{2}{*}{1.85×} \\ \cline{3-4}
                      &                           & Turbo    & 85.3/\textbf{84.9} &                        \\ \cline{2-5} 
                      & \multirow{2}{*}{ViT-L/16} & ToMe     & 88.1/87.7 & \multirow{2}{*}{1.98×} \\ \cline{3-4}
                      &                           & Turbo    & 88.1/\textbf{87.9} &                        \\ \cline{2-5} 
                      & \multirow{2}{*}{ViT-H/14} & ToMe     & 88.6/88.2 & \multirow{2}{*}{1.91×} \\ \cline{3-4}
                      &                           & Turbo    & 88.6/\textbf{88.4} &                        \\ 
\bottomrule
\end{tabular}
\vspace{-0.2cm}
\caption{\textbf{Uni-Modal ImageNet Classification Task.} Turbo shows better results on all-size models, providing similar acceleration.
} 
\label{tab:uni}
\vspace{-0.2cm}
\end{table}

\subsection{Ablation Study \& Discussion} \label{sec:abla}
We here make comprehensive ablations to dissect components. Without loss of generality, all analysis below are conducted on the BLIP image captioning task.

\vspace{0.1cm}
\noindent \textbf{Effectiveness of Mutual Redundancy \& Semantic Value.} 
Mutual redundancy serves as one key component of our progressive merging strategy. As illustrated in Section~\ref{sec:theory}, mutual redundancy reveals the hidden dependency relationship between tokens. Table~\ref{tab:degree} demonstrate the performance boost w/ and w/o mutual redundancy. By adding mutual redundancy solely or on the basis of semantic value, we can witness an obvious boost in final performance. We also study the effect of semantic value in Table~\ref{tab:degree}. Similarly, semantic value also has an apparent enhancement for model performance. By balancing these two components, we launch the information degree to achieve the best results.

\begin{figure}[t]
\begin{center}
\includegraphics[width=0.475\textwidth] {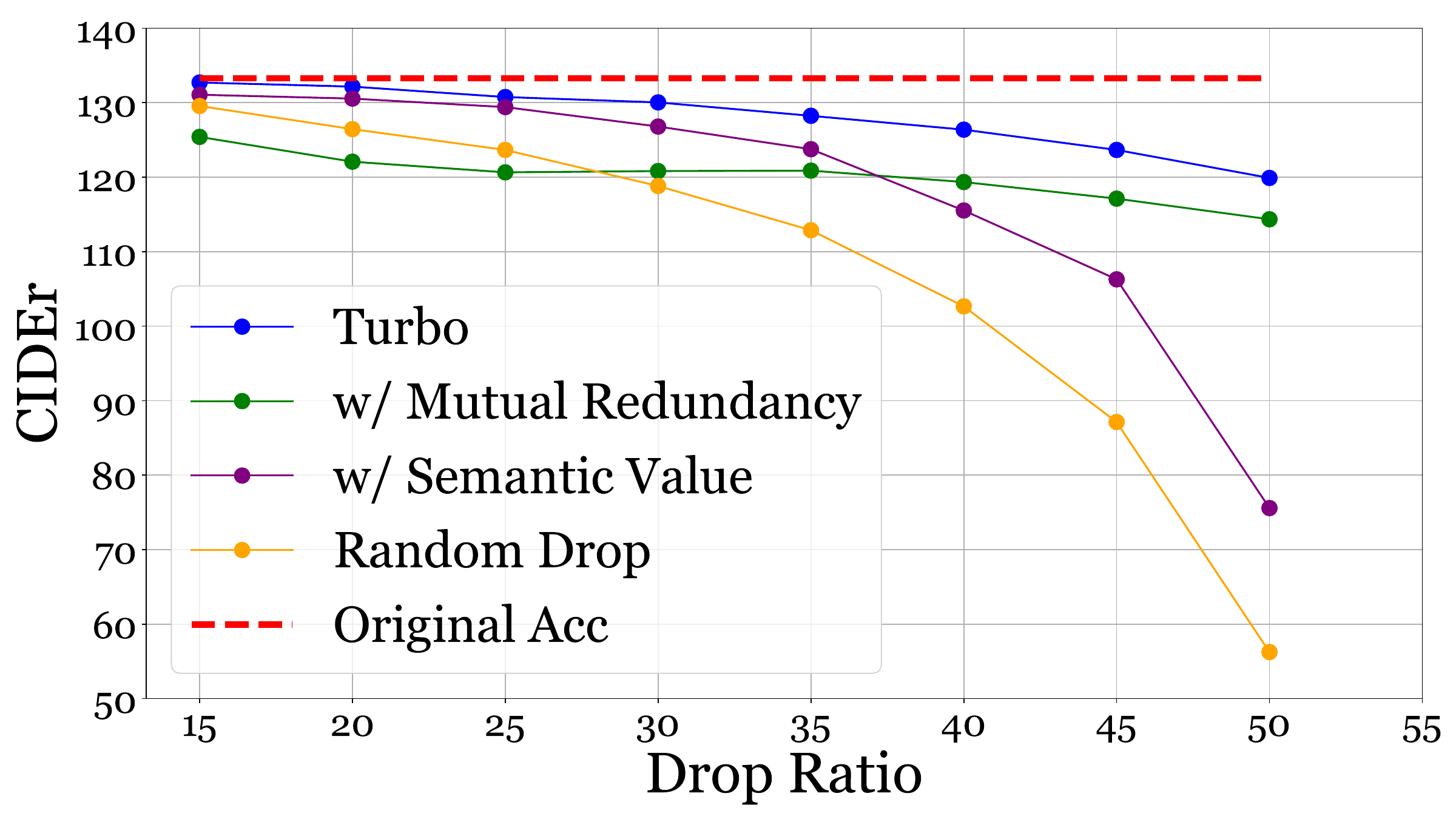}
\end{center}
\vspace{-0.7cm}
\caption{\textbf{Ablation Study on Drop Ratio $\Upsilon$.} We evaluate separately the effectiveness of key components on various $\Upsilon$. Semantic value retains superior performance when $\Upsilon$ is small while mutual redundancy possesses better stability on large $\Upsilon$. By combining these two drop strategies, our method achieves both competitive performance and stability on the whole scope.}
\label{fig:abla}
\end{figure}

\vspace{0.1cm}
\noindent \textbf{Trend of Key Components with Respect to Drop Ratio.} 
In addition to the ablation study on one single drop ratio $\Upsilon$, we investigate the fluctuation of performance in a wide range of $\Upsilon$. In figure~\ref{fig:abla}, we demonstrate the performance of each component combination on different drop ratios. On one hand, the scheme with only mutual redundancy is relatively stable on the large scope but performs badly on small $\Upsilon$. This accords with our prediction in Section~\ref{sec:theory} that mutual redundancy can't distinct important tokens. On account of the aggregation of semantic-rich tokens at early stages, it will inevitably lose fine-grained information. Some visualization results will be presented in supplementary materials to demonstrate this phenomenon.

On the other hand, scheme that is solely guided by semantic values possesses satisfying results on small $\Upsilon$ but drops dramatically on large $\Upsilon$. This is due to the fact that the approximation of information degree in expression~\ref{eq:12} only holds when the neighbourhood condition is satisfied. When drop ratio gets larger, the neighbourhood condition is no longer valid, resulting in a dramatic drop in performance. By combing semantic value with mutual redundancy, we achieve superior performance on all the scope of $\Upsilon$.

\begin{table}[t]
\footnotesize
\setlength\tabcolsep{3.5pt}
\centering
\begin{tabular}{c|ccc|ccc}
\toprule 
& Redundancy & Semantics & Fusion & B@4 & CIDEr & Throughput \\ 
\hline \hline
A1 &   &  & - & 34.5 & 112.8 &  73.6 \\
A2 & \checkmark  &  & - & 35.5 & 120.9 & 70.3 \\
A3 &  & \checkmark & - & 36.4 & 123.8 & 67.9 \\
A4 & \checkmark & \checkmark  & $+$ & 38.2 & 130.0 & 67.6 \\
A5 & \checkmark & \checkmark  & $\times$ & 38.2 & 129.9 & 62.8 \\
\bottomrule
\end{tabular}
\vspace{-0.1cm}
\caption{\textbf{Ablation Study on Key Components.} We validate the effectiveness of each component, and compare fusion strategies.}
\vspace{0.3cm}
\label{tab:degree}
\end{table}

\vspace{0.1cm}
\noindent \textbf{Fusion for Information Degree.} We mainly adopt two fusion strategies illustrated in Eq.~\ref{eq:balance}. As shown in Table~\ref{tab:degree}, the coupled product strategy and weighted averaging strategy achieves similar performance, further proving the robustness of the turbo module. However,due to the complexity of multiplication over addition, the coupled product strategy is slower than weighted averaging strategy. Furthermore, several complex fusion strategies {\em i.e.}, dynamic $\alpha$ on different layers are also studied. However, they all result in minor enhancement, therefore we here simply adopt the weighted averaging for efficient merging.

\begin{figure}[t]
\begin{center}
\includegraphics[width=0.475\textwidth] {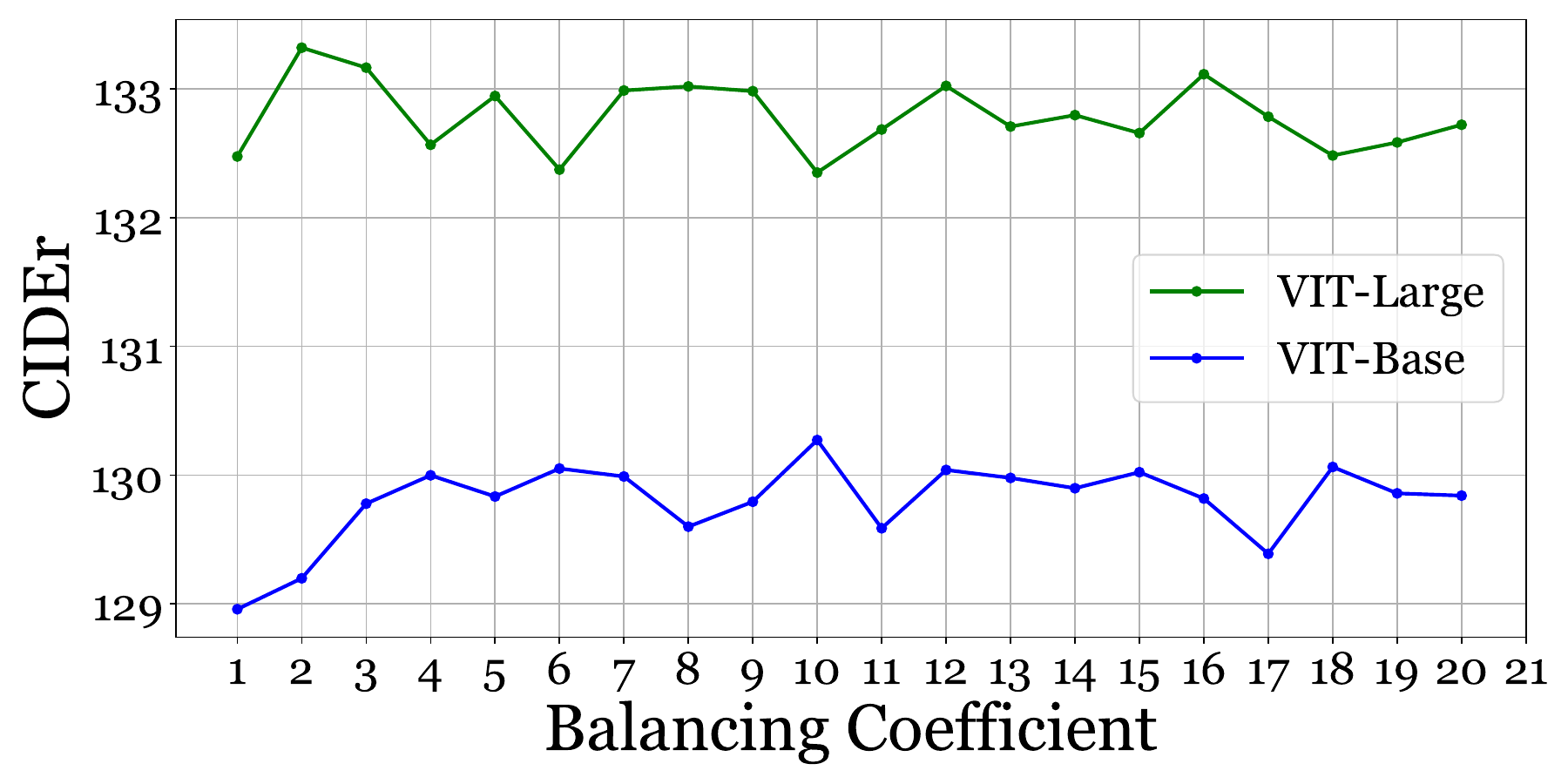}
\end{center}
\vspace{-0.4cm}
\caption{\textbf{Ablation Study of Balancing Coefficient.} We evaluate the influence of balancing coefficient on image captioning task using BLIP VIT-Base and VIT-Large models. The result proves the robustness of our Turbo, as the performance varies slightly in a certain range.}
\label{fig:robust}
\end{figure}

\vspace{0.1cm}
\noindent \textbf{Framework Robustness.} The balancing coefficient $\alpha$ in Eq.~\ref{eq:balance} is a critical hyper-parameter, which can be toxic for the framework, since we have to take time to carefully choose $\alpha$ if it has a vital influence on final performance. To this end, we conduct an experiment on the choice of $\alpha$. We test the impact of $\alpha$ on image captioning for BLIP (VIT-Base and VIT-Large). Figure~\ref{fig:robust} reports the results. From the range $\alpha \in \{1,2,...,20\}$, the performance fluctuates slightly on both models with different sizes, indicating that the framework is unaware of the balancing coefficient $\alpha$ within a certain range. This proves our robustness.

\begin{figure}[t]
\begin{center}
\includegraphics[width=0.48\textwidth] {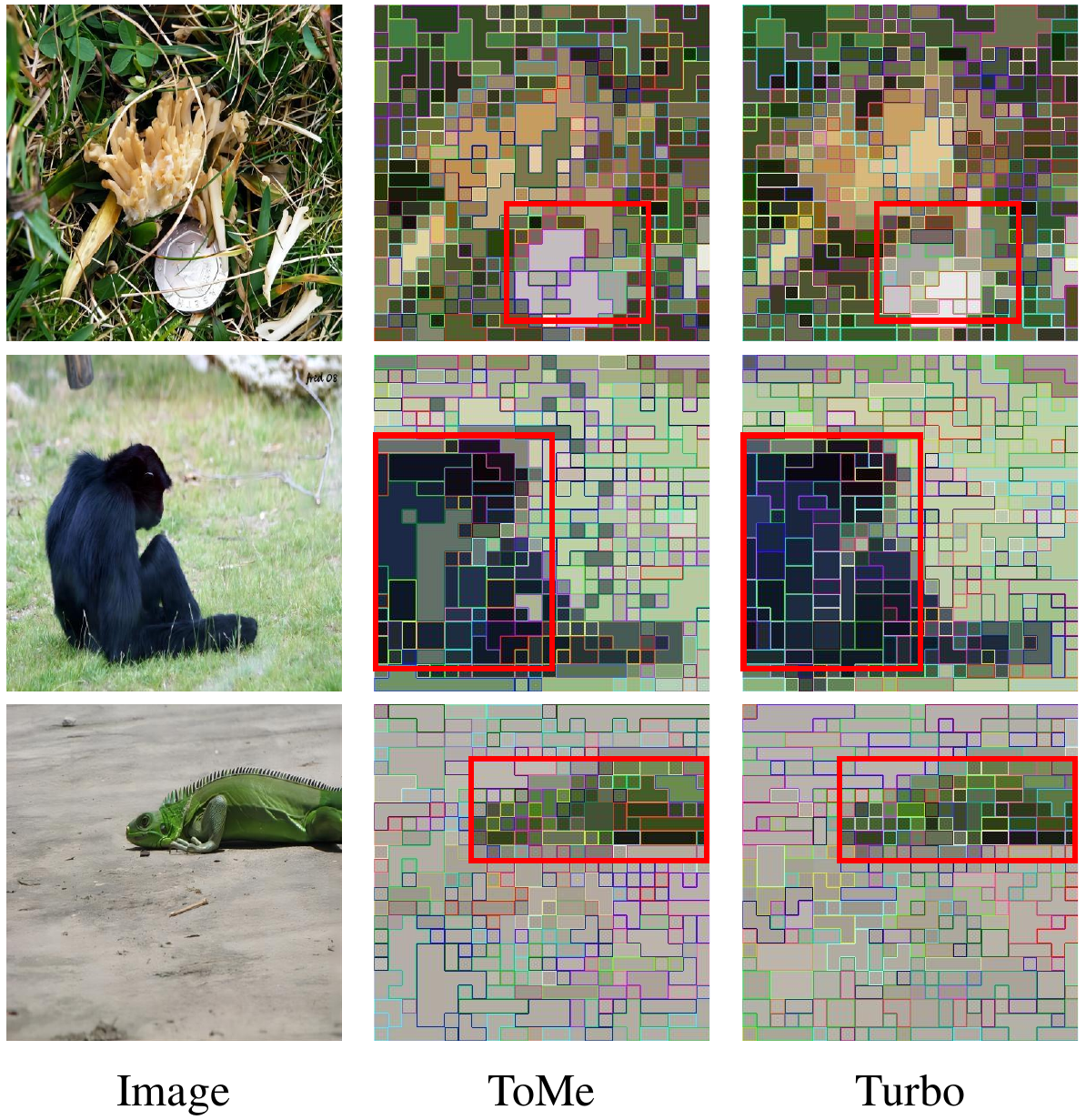}
\end{center}
\vspace{-0.5cm}
\caption{\textbf{Visualizations of Token Merge Process.} Turbo tends to merge background patches while retaining foreground patches with semantic value, thus preserving more critical information.}
\label{fig:mergepro}
\end{figure}

\subsection{Visualization of Merging Process}
As shown in Figure~\ref{fig:mergepro}, we analyze the process of merging tokens under understanding tasks (image classification). And ToMe~\cite{bolya2022token} is selected for comparisons. In each image, we highlight the foreground object in the red box. 
There is one clear trend that, comparing to ToMe, our Turbo merges more background patches, while retaining most foreground patches with semantic value. 
As a result, the data accelerated by Turbo will maintain more comprehensive and fine-grained information, ensuring higher task performance, especially on difficult tasks like image captioning and vqa.

\section{Conclusion}
This paper proposes Turbo plug-in for VLMs' acceleration, from one novel perspective of data de-redundancy. In pursuit of speed-performance trade-offs, Turbo defines information degree for data reduction, taking into account two critical components: mutual redundancy and semantic value. The former evaluates data overlap between sequential tokens; the latter evaluates token's contribution to overall semantics. By eliminating data redundancy from the source, Turbo generally compatible to various VLMs with trivial development efforts. Extensive experiments and thorough ablations show the significance of Turbo, across multi-modality/uni-modality and understanding/generation.

% \clearpage

{\small
\bibliographystyle{ieee_fullname}
\bibliography{egbib}
}
\clearpage

In the supplementary material, we first provide more details about our plug-in turbo module to facilitate reproduction. Then more ablation studies on different fusion strategies are provided. Finally we demonstrate some visualization results to justify our arguments, including the generated images from stable diffusion and intermediate merging process of understanding tasks.

\section{Implementation Details}
\subsection{Turbo Architecture for Understanding Tasks}
Turbo can be easily plugged in almost any pre-trained, attention-based models to reduce the total sequence length block by block, with no need for further training or adjustment. In practice, we replace all the attention blocks by the turbo module.

For Turbo, we merge tokens progressively based on their information degree, which is defined as follows:
\begin{equation} 
    {\mathcal{E} = \mathcal{R}-\alpha\mathcal{I}, \quad \mathcal{E} = \frac{\mathcal{R}}{\mathcal{I}}.
    }
\end{equation} 

Information degree takes both mutual redundancy and semantic values into account, encouraging insignificant, similar tokens to be merged preferentially. This merging strategy reduces the amount of tokens with duplicated or low informativity, compressing the token sequence with minor information loss. Inspired by~\cite{ahmed2017diverse,bolya2022token}, we follow the bipartite soft matching to calculate mutual redundancy and apply the merging strategy to aggregate tokens. Specifically, we leverage keys (K) or queries (Q) in QKV attention and the cosine similarity metric between tokens to measure the similarity between tokens. We define mutual redundancy between two tokens as the similarity between them. By adding the quantity of semantic value, which is the attention proportion of each token, the information degree is obtained. 

To avoid excessive computational cost of calculating similarity matrix of the whole token sequence, we leverage bipartite soft matching to speed up the merging process. Suppose the drop ratio is $\Upsilon$, which means we will reduce the amount of tokens by number $\Upsilon$ in each block. In every block, we divide the tokens into two partitions $B$ and $C$ of the same size (if the number of tokens is odd, one partition has 1 more tokens than the other). Then we calculate the mutual redundancy between the two partitions $B$ and $C$. After adding the semantic value of each token on partition $B$, we sort the information degree of $B$ and merge the top $\Upsilon$ tokens onto $C$. At last we concatenate the sequence back to continue the forwarding process. In this way, we reduce the length of token sequence by $\Upsilon$ in each block. Notice that We call $\Upsilon$ the drop ratio, but it is in fact the number of tokens we reduce every attention block, which is a real ratio by dividing the length of the token sequence. We also note that the semantic value are naturally contained in visual $\mathrm{cls}$ self-attention map or texture $\mathrm{cls}$ cross-attention map depending on the model structure.

When the merging process are finished, some tokens can represent several different patches. This can
change the outcome of softmax attention and thus influence the semantic value. So We fix this with a minor modification:
\begin{equation}
\mathbf{A} = \text{softmax} \left( \frac{QK^T}{\sqrt{d}} + \log s \right),
\end{equation}
where $s$ is the number of patches/tokens represented by the merged tokens all along the merging procedure.

In order to avoid excessive merging, {\em i.e.}, merging too many tokens and leave only a few tokens in final blocks. This will cause insufficient expression ability problem and we observe a sharp performance drop at certain drop ratio. So we set up a threshold $r$ for the least number of tokens in the final stage, to mitigate the dramatic drop on large $\Upsilon$. Table \ref{tab:555} shows the effectiveness of this restriction for preventing a sudden performance decline.

\subsection{Turbo Architecture for Generation Tasks}
For low-level generative tasks, Stable Diffusion~\cite{rombach2022high} is one popular backbone. Here, Turbo contains one merging module and one inverse-sampling module. For the merging module, we attach Turbo acceleration on UNet of Stable Diffusion, as UNet consumes the most computation. More specifically, UNet usually consists of three key components: self-attention, cross-attention and FFN. We add Turbo merging/restoring before/ after each component. For self-attention and FFN, Turbo merging is calculated by one visual modality; while for cross-attention, Turbo merging is calculated by visual-textual modalities.

We evaluate by generating $2000$ images, each resolution is $512*512$. The text classes used are from ImageNet-1k. We use FID scores to metric the generation quality.

\section{More Experiments}
\begin{table}[t]
\footnotesize
\setlength\tabcolsep{5pt}
\centering
\begin{tabular}{c|ccc|ccc}
\toprule 
Strategy & $\alpha$ & $\beta$ & $\gamma$ & B@4 & CIDEr & Throughput \\ 
\hline \hline
S1 & 6 & - & - & 38.2 & 130.0 & 67.6 \\
S2 & -  & - & - & 38.2 & 129.9 & 62.8 \\
S3 & 1 & - & - & 37.8 & 128.5 & 61.9\\
S3 & 2 & -  & - & 37.9 & 128.4 & 61.9 \\
S3 & 3 & -  & - & 36.9 & 125.7 & 61.9 \\
S3 & 4 & -  & - & 36.1 & 123.1 & 61.9 \\
S4 & 6 & 0.9 & 1 & 38.2 & 129.5 & 67.4\\
S4 & 6 & 0.9  & 2 & 38.3 & 129.5 & 67.4\\
S4 & 6 & 0.9  & 3 & 38.1 & 129.5 & 67.4\\
S4 & 6 & 0.9  & 4 & 38.2 & 129.5 & 67.4\\
S4 & 6 & 0.9  & 5 & 38.1 & 129.7 & 67.4\\
S4 & 6 & 0.9 & 6 & 38.0 & 129.2 & 67.4\\
S4 & 6 & 0.9  & 7 & 38.0 & 129.5 & 67.4\\
S4 & 6 & 0.9  & 8 & 38.4 & 130.2 & 67.4\\
S4 & 6 & 0.9  & 9 & 38.2 & 129.9 & 67.4\\
S4 & 6 & 0.9  & 10 & 38.2 & 129.7 & 67.4\\
S4 & 6 & 1.1 & 1 & 38.3 & 130.0 & 67.4\\
S4 & 6 & 1.1  & 2 & 38.3 & 130.3 & 67.4\\
S4 & 6 & 1.1  & 3 & 38.4 & 130.1 & 67.4\\
S4 & 6 & 1.1  & 4 & 38.3 & 129.9 & 67.4\\
S4 & 6 & 1.1  & 5 & 38.5 & 129.9 & 67.4\\
S4 & 6 & 1.1 & 6 & 38.1 & 129.3 & 67.4\\
S4 & 6 & 1.1  & 7 & 38.2 & 129.6 & 67.4\\
S4 & 6 & 1.1  & 8 & 38.5 & 130.3 & 67.4\\
S4 & 6 & 1.1  & 9 & 38.2 & 129.8 & 67.4\\
S4 & 6 & 1.1  & 10 & 38.2 & 129.6 & 67.4\\
\bottomrule
\end{tabular}
\vspace{-0.1cm}
\caption{\textbf{Ablation Study on Fusion Strategies.} We adopt several fusion strategies and test their performance under different coefficients. Some complex fusion strategies can achieve slightly better results, but in order to keep the form easy to apply, we adopt the simple weighted average in other experiments.}
\label{tab:strategy}
\end{table}
\vspace{-0.1cm}
\subsection{Ablation Study on Fusion Strategy}
We design 4 different fusion strategies to combine mutual redundancy ($\mathcal{R}$) with semantic value ($\mathcal{I}$):

\begin{align} 
   & \mathrm{S1}: \mathcal{E} = \mathcal{R}-\alpha\mathcal{I}, \\
   & \mathrm{S2}: \mathcal{E} = \frac{\mathcal{R}}{\mathcal{I}}, \\
   & \mathrm{S3}: \mathcal{E} = \mathcal{R}(1-\alpha\mathcal{I}), \\
   & \mathrm{S4}: \mathcal{E} = \mathcal{R}-\beta^{\left| \gamma-\mathrm{block\_id}\right|}\alpha\mathcal{I},
\end{align} 
where $\mathrm{S4}$ is designed to allow dynamic scales between $\mathcal{R}$ and $\mathcal{I}$ on different blocks. For example, if $\beta>1$ and $\gamma=6$, then the scale of $\mathcal{I}$ will reach its maximum on block 6 and attain the minimum value on two-end blocks. 

As shown in table~\ref{tab:strategy}, we have done extensive experiments on the four fusion strategies with different coefficients. Though $\mathrm{S4}$ attains the best result, due to its complexity and such slight performance gain, {\em i.e.}, three hyper-parameters to be determined with a gain of only 0.3 on CIDEr, we thus adopt the simple weighted average fusion strategy ($\mathrm{S1}$) on our turbo module.

\subsection{Ablation Study on Threshold}
\begin{table}[t]
\footnotesize
\setlength\tabcolsep{5pt}
\centering
\begin{tabular}{cc|ccc}
\toprule 
$\Upsilon$ & Threshold & B@4 & CIDEr & Throughput \\ 
\hline \hline
65 & 0 & 28.8 & 95.5 &  107.6 \\
65 & 70 & 31.1 & 104.3 & 103.3 \\
65 & 130 & 34.2 & 115.3 & 100.8 \\
70 & 0 & 27.3 & 89.4 & 113.0 \\
70 & 70 & 31.0 & 103.7 & 110.7 \\
\bottomrule
\end{tabular}
\vspace{-0.1cm}
\caption{\textbf{Ablation Study on Threshold.} We validate the effectiveness of threshold to prevent dramatic performance drop on large $\Upsilon$. With slight decline on speed, model performance with threshold on minimum token length can maintain a smoother decrease when $\Upsilon$ getting larger.}
\label{tab:555}
\end{table}
When applying large drop ratios $\Upsilon$ on turbo module, we witness a sharp drop of model performance. We argue that this phenomenon is due to the insufficient expression ability of token sequence length below a certain threshold. So we append a threshold on minimum number of tokens left in the final block. Specifically, we stop the token merging process once the token sequence length is below the threshold. Results in table~\ref{tab:555} have demonstrate the effectiveness of such a threshold on large $\Upsilon$, improving the model performance by over 10 points on CIDEr with slight acceleration declines.

\end{document}